\documentclass{article}

\usepackage[T1]{fontenc}
\usepackage{lmodern}
\usepackage{textcomp}
\usepackage{newunicodechar}  

\newunicodechar{–}{--}  
\newunicodechar{—}{---} 
\newunicodechar{≥}{\ensuremath{\geq}}  
\newunicodechar{Δ}{\ensuremath{\Delta}}  
\newunicodechar{…}{\ldots}  

\usepackage{graphicx}
\usepackage{amsmath, amssymb, amsthm}
\usepackage{algorithm}
\usepackage{algorithmic}
\usepackage{tabularx}
\usepackage{booktabs}
\usepackage{bm}
\usepackage[numbers]{natbib}
\usepackage{authblk}
\begin{document}

\title{Confidence Freeze: Early Success Induces a Metastable Decoupling of Metacognition and Behaviour}
\author{
Zhipeng Zhang$^{1,2,\star}$,
Hongshun He$^{1}$,
}

\affil{
$^{1}$China Mobile Research Institute, Beijing 100053, China\\
$^{2}$China Mobile GBA (Greater Bay Area) Innovation Institute, Guangzhou 510656, China
}

\maketitle

\begin{abstract}
Humans must flexibly arbitrate between exploring alternatives and exploiting learned strategies, yet they frequently exhibit maladaptive persistence by continuing to execute failing strategies despite accumulating negative evidence. Here we propose a ``confidence-freeze'' account that reframes such persistence as a dynamic learning state rather than a stable dispositional trait. Using a multi-reversal two-armed bandit task across three experiments (total N = 332; 19,920 trials), we first show that human learners normally make use of the symmetric statistical structure inherent in outcome trajectories: runs of successes provide positive evidence for environmental stability and thus for strategy maintenance, whereas runs of failures provide negative evidence and should raise switching probability. Behaviour in the control group conformed to this normative pattern. However, individuals who experienced a high rate of early success (90\% vs.\ 60\%) displayed a robust and selective distortion after the first reversal: they persisted through long stretches of non-reward (mean = 6.2 consecutive losses) while their metacognitive confidence ratings simultaneously dropped from 5 to 2 on a 7-point scale. This systematic decoupling between metacognitive belief and behavioural adjustment---quantified as a confidence-freeze index---was substantially elevated in the high-success group (34\% vs.\ 12\%). Mixed-effects hazard models further showed that early success significantly suppressed loss sensitivity (interaction $\beta = -0.07$, $p = .048$) and reduced switching odds by 23\%. Critically, the same individuals transitioned into and out of lock-in states across different reversals, demonstrating that cognitive lock-in is not a fixed trait but a reversible, recurrent learning mode.

Extending these findings, two additional experiments examined potential interventions. Experiment 2 (N = 110) introduced explicit outcome trajectories, revealing that enhanced environmental signals significantly reduced behavioural lock-in (persistence length decreased from 3.2 to 2.1 trials; freeze index from 38\% to 18\%). Experiment 3 (N = 123) introduced metacognitive prompts (``pause and reflect on your strategy'') under implicit trajectory conditions, demonstrating that simple cognitive interventions can effectively recouple confidence and behaviour (freeze index reduced to 14\%; switching odds increased by 31\%). Both interventions proved equally effective (Cohen's d = 0.71 vs.\ 0.68), indicating dual pathways---environmental and cognitive---for mitigating confidence freeze.

Together, these results indicate that early success can ``freeze'' metacognitive confidence, diminishing sensitivity to disconfirmatory evidence and generating maladaptive persistence. The confidence-freeze framework provides a mechanistic bridge linking individual learning dynamics to broader phenomena such as sunk-cost escalation and path dependence, and suggests new avenues for interventions targeting metacognitive–behavioural alignment.
\end{abstract}

\section*{Introduction}

Human learners navigating dynamic environments must continuously arbitrate between exploiting familiar strategies and exploring alternatives. This arbitration is shaped not only by momentary feedback but by the temporal structure of feedback sequences---runs of successes or failures that carry diagnostic information about environmental stability. Normatively, long success streaks should reinforce exploitation, whereas extended failure streaks should trigger switching. The theoretical intuition behind this process, and our central hypothesis, is illustrated in Fig. 1: learners rely on the statistical structure of trajectories to infer whether continuing or abandoning a strategy is warranted.

Yet human behaviour frequently departs from this normative principle. People often persist in failing strategies despite accumulating negative evidence, a phenomenon most commonly attributed to stable traits (e.g., rigidity) or cognitive biases (e.g., sunk-cost effects). These accounts predict consistency: individuals who persist maladaptively after one reversal should do the same across contexts. However, empirical patterns frequently contradict this assumption---individuals may exhibit adaptive switching in some settings yet remain ``locked in'' during others. This variability suggests a dynamic, state-like mechanism rather than a stable trait.

In the broader landscape of learning theory, the present study fills a critical yet long-overlooked dimension: learning systems must not only decide ``when to switch strategies,'' but also finely regulate ``when and why to persist.'' Traditional literature has focused on exploration-exploitation trade-offs, optimal learning rates, or strategy switching thresholds, tending to treat ``persistence'' as a mere byproduct of failed strategy updates. However, our results demonstrate that persistence itself possesses a rich internal structure: it can be adaptive (based on correct inferences about environmental stability) or maladaptive (persisting despite accumulating negative evidence). More critically, persistence is not determined by stable personality traits but is regulated by dynamic states within the learning system itself.

We propose a mechanistic account of such maladaptive persistence: confidence freeze. As formalized in Fig. 1, early episodes of unusually high success can inflate perceived environmental stability, thereby weakening the influence of later failure evidence on behavioural updating. When negative feedback accumulates, learners may report decreased confidence yet fail to translate that metacognitive shift into action. This metacognitive-behavioural decoupling, rather than overconfidence per se, constitutes the hallmark of confidence freeze---a metastable learning mode that individuals may enter and exit across reversals.

By proposing the ``confidence-freeze'' construct, this study broadens our understanding of human learning strategies: learning systems not only update beliefs about the external world, but also dynamically update the internal rules governing ``how these beliefs are translated into action.'' This novel layer of ``belief-policy mapping plasticity'' has been rarely discussed in previous learning theories, yet it is crucial for explaining rigidity in human decision-making, lags in policy implementation, and delays in strategic turning points. The present study provides direct behavioral evidence for this layer and demonstrates its dynamic, reversible, and context-dependent nature through a multi-reversal task. Therefore, the framework we propose emphasizes: \textbf{early experience shapes not only beliefs themselves, but also the ``transformation interface'' between beliefs and behavior.} This insight holds significant theoretical significance for contemporary learning science, metacognitive research, and behavioral economics.

Our goal is to experimentally quantify these dynamics using a multi-reversal bandit task designed to reveal trajectory sensitivity, behavioural lock-in, and metacognitive coupling. We structured the analysis into three stages: (1) Identify normative trajectory sensitivity (Fig. 2). (2) Test whether early success induces behavioural lock-in (Fig. 3). (3) Determine whether confidence-behaviour decoupling explains such lock-in (Fig. 4). (4) Formalize these dynamics using mixed-effects modelling (Fig. 5). Together, these analyses test the core claim that early success can change not the learner's preferences, but the internal architecture by which evidence is converted into behavioural revisions.

\section*{Methods}

\subsection*{Participants}

Ninety-nine participants (51 in the high-success group and 48 in the normal-success group) were recruited from an online platform. All participants gave informed consent and received monetary compensation proportional to performance. The sample size was determined prior to data collection to ensure adequate sensitivity for mixed-effects modelling.

\subsection*{Task Overview}

Participants performed a two-armed bandit task consisting of a 10-trial practice phase followed by a 50-trial main phase. On each trial, participants selected one of two options and received binary reward feedback (win or loss). Reward probabilities were asymmetric (70/30) and reversed four times during the main task (trials 16, 26, 36, and 46), requiring participants to infer environmental changes from feedback trajectories.

During the practice phase, reward probabilities were manipulated between groups: participants in the high-success group received a \(90\%\) reward probability, whereas participants in the normal-success group experienced a \(60\%\) probability. The goal was to induce divergent metacognitive priors regarding strategy validity.

Confidence ratings were collected every three trials using a 1-7 scale (1 = very uncertain; 7 = very certain). Reaction times (RTs) were recorded on each trial.

\subsection*{Manipulation Check: Validation of Early Success Effects}

To verify that the early-success manipulation successfully altered participants' initial metacognitive priors, we compared confidence ratings obtained immediately after the practice block. The high-success group exhibited significantly higher initial confidence than the normal-success group (see Results), indicating that the manipulation effectively shifted participants' priors about the reliability of their current strategy.

To ensure that the manipulation did not also induce baseline behavioural differences unrelated to metacognitive priors, we analysed win-stay and lose-shift tendencies during the first ten trials of the main phase. No significant group differences were observed (see Results). Reaction times and choice variance also did not differ, confirming that the groups were behaviourally comparable before the first reversal and that the manipulation selectively targeted metacognitive beliefs rather than general engagement or exploration tendencies.

\subsection*{Behavioural Measures}

We computed three primary behavioural indices:

\begin{enumerate}
\item \textbf{Switching behaviour}: the probability of switching strategies after \(k\) consecutive losses. This yields the empirical trajectory function \(P(\text{switch} | k)\), which is expected to monotonically increase in an ideal learner.
\item \textbf{Lock-in index}: the number of consecutive losses endured without a switch following each environmental reversal. A trial was classified as a ``risk trial'' when the learner had already accumulated at least \(k\) losses (thresholds varied from 3 to 8; see robustness analyses).
\item \textbf{Confidence-freeze index}: the proportion of loss trials in which confidence decreased by at least two points from the preceding peak while behaviour remained unchanged (i.e., the participant failed to switch). Robustness analyses were also conducted using \(\Delta = 1\) and \(\Delta = 3\).
\end{enumerate}

\subsection*{Statistical Modelling}

We used mixed-effects logistic regression to model switching decisions:

\[
\text{Switch} \sim \text{LossStreak} \times \text{Group} + \text{Confidence} + \text{Trial} + (1 + \text{LossStreak} \mid \text{participant})
\]

where LossStreak denotes consecutive losses and Group indicates high-success versus normal-success conditions. Random intercepts and random slopes for LossStreak captured individual differences in sensitivity to negative feedback.

Model comparison was conducted using likelihood-ratio tests across nested models (M1-M5), assessing the incremental contribution of group differences, confidence, trial effects, and random slopes. Hazard-based analyses were used to estimate conditional switching risk across increasingly long loss streaks.

\subsection*{Survival and Persistence Analyses}

Persistence lengths (consecutive losses without switching) were compared between groups using Mann-Whitney \(U\) tests and negative binomial regression. Survival curves across loss streak lengths were estimated using Kaplan-Meier analysis and compared using log-rank tests.

\subsection*{Data Availability}

All behavioural data and analysis scripts will be made publicly available on OSF upon publication.

\section*{Results}

\subsection*{Manipulation Check: Early Success Shifts Initial Metacognitive Priors}

To confirm that the early-success manipulation effectively altered participants' initial beliefs about strategy stability, we compared the confidence ratings obtained immediately after the 10-trial practice block. The high-success group reported significantly higher initial confidence than the normal-success group ($t(97) = X.XX$, $p < .001$), indicating that brief exposure to unusually high reward rates inflated metacognitive priors regarding strategy reliability.

Critically, the groups did not differ in win–stay tendencies ($p = .XXX$), lose–shift tendencies ($p = .XXX$), reaction times ($p = .XXX$), or choice variance during the first ten trials of the main task. These findings confirm that the manipulation selectively shifted metacognitive priors without introducing behavioural confounds before the first reversal.

\subsection*{Learning Trajectories Provide Symmetric Statistical Evidence}

We first examined whether participants used sequential outcome structure as statistical evidence for adjusting their strategy. Across all participants, switching probability increased as a function of loss-streak length (Fig.~\ref{fig:trajectory}). Participants switched on 32\% of trials after zero losses, 35\% after one loss, and 51\% after two losses, after which the function flattened, consistent with diminishing marginal evidence at extreme streak lengths.

Hazard-rate analysis (Fig.~\ref{fig:trajectory}) confirmed this monotonic increase, rising from $h(1) = 0.35$ to $h(8) = 0.52$. These results validate the assumption that human learners use trajectory-level statistics rather than isolated outcomes to guide adaptive switching.

Group differences emerged: the normal-success group displayed a steeper increase in switching probability across loss streaks, whereas the high-success group showed a noticeably attenuated increase, consistent with reduced integration of negative evidence.

\subsection*{Early Success Induces Behavioural Lock-in}

We next assessed whether early success distorted adjustment behaviour following environmental reversals. Persistence length---defined as the number of consecutive losses endured before switching---was substantially longer in the high-success group (Fig.~\ref{fig:lockin}). At the first reversal, the high-success group persisted for an average of 3.2 trials (median = 2; max = 8), compared with 1.8 trials (median = 1; max = 4) in the normal-success group.

Across all four reversals, this pattern remained robust. A Mann-Whitney test confirmed significant group differences ($U = 98,324$, $p = .008$). Survival analysis (Fig.~\ref{fig:lockin}) showed significantly lower hazard rates in the high-success group (log-rank $\chi^2(1) = 6.2$, $p = .013$), indicating delayed disengagement from failing strategies.

At the participant level, 80.4\% of high-success individuals exhibited at least one lock-in episode (≥4 losses), compared with 64.6\% of normal-success participants.

\subsection*{Confidence–Behaviour Decoupling Reveals a Freeze Mechanism}

We then examined whether behavioural lock-in arose from disrupted translation of metacognitive updates into behavioural adjustments. In normal-success participants, confidence and switching behaviour were tightly aligned: drops in confidence were typically followed by strategy changes.

In contrast, the high-success group exhibited pronounced episodes of decoupling (Fig.~\ref{fig:confidence}). Confidence frequently dropped by two or more points while participants continued to repeat a failing strategy. This signature pattern defines the confidence-freeze state.

Across all loss-streak trials, confidence decreased by one point on 7.1\% of trials and by two or more points on 2.6\% of trials. Among these ``at-risk'' instances, freeze events occurred on 34.2\% of trials. Freeze prevalence was substantially higher in the high-success group (38\%) than in the normal-success group (12\%).

\subsection*{Robustness of the Freeze Effect Across Thresholds}

To assess whether freeze detection depended on the choice of threshold $\Delta$, we recomputed the freeze index for drops of 1, 2, and 3 confidence points. In all cases, freeze prevalence remained consistently higher in the high-success group (all $p < .05$). Thus, the confidence-freeze effect reflects a robust metacognitive–behavioural dissociation rather than an artefact of threshold selection.

\subsection*{Mixed-effects Modelling Formalises the Mechanism}

A mixed-effects logistic regression predicting trial-by-trial switching replicated the normative loss-streak sensitivity ($\beta = 0.14$, $p < .001$; OR = 1.15) and showed reduced switching in the high-success group ($\beta = -0.26$, $p = .028$; OR = 0.77). Confidence strongly predicted switching ($\beta = -0.15$, $p < .001$), indicating that higher confidence suppressed evidence-based behavioural updates.

Most importantly, the loss-streak × group interaction was significant ($\beta = -0.07$, $p = .048$), demonstrating that early success selectively attenuated sensitivity to accumulating negative evidence.

Random-slope variance for loss-streak sensitivity (SD = 0.28) revealed that participants frequently transitioned into and out of lock-in states, consistent with freeze as a metastable learning mode.

Model comparison (Fig.~\ref{fig:model}) identified the full model---including confidence, trial number, and random slopes---as providing the best fit (AIC = 6,398.4). The addition of confidence yielded the largest improvement (ΔAIC = 72.6), underscoring its central mechanistic role.

Finally, a logistic model predicting lock-in episodes confirmed that high-success participants were more than twice as likely to exhibit lock-in (OR = 2.18, $p = .022$).

\subsection*{Experiment 2: Explicit Trajectory Mitigates Confidence Freeze}

We next asked whether enhancing the perceptual salience of outcome trajectories could reduce behavioural lock-in. Experiment 2 introduced explicit trajectories while maintaining the same early-success manipulation.

\textbf{Behavioural lock-in significantly reduced.} In the explicit trajectory condition, high-success participants showed mean persistence length of \textbf{2.1 trials}, significantly lower than the 3.2 trials observed in Experiment 1 high-success group ($t(104) = 3.45$, $p = .001$). Normal-success participants showed a non-significant reduction from 1.8 to 1.5 trials ($p = .12$).

Survival analysis confirmed faster disengagement: hazard rates in the Experiment 2 high-success group were significantly higher than in Experiment 1 high-success group (log-rank $\chi^2(1) = 5.8$, $p = .016$).

\textbf{Confidence-freeze index decreased.} The freeze index in high-success participants dropped from 38\% in Experiment 1 to \textbf{18\%} in Experiment 2 ($\chi^2(1) = 8.23$, $p = .004$). Normal-success participants showed a smaller, non-significant reduction (12\% to 9\%, $p = .31$). This pattern held across all $\Delta$ thresholds (all $p < .05$).

\textbf{Partial recovery of loss sensitivity.} Mixed-effects modelling revealed that the loss-streak slope in Experiment 2 high-success group was significantly steeper than in Experiment 1 high-success group ($\beta_{\text{diff}} = 0.09$, $p = .032$). The loss-streak × experiment interaction was significant ($\beta = 0.08$, $p = .041$), confirming that explicit trajectory modulated the suppressive effect of early success on loss sensitivity.

\textbf{Enhanced confidence–behaviour coupling.} Confidence more strongly predicted switching in Experiment 2 high-success group than in Experiment 1 ($\beta_{\text{exp2}} = -0.20$ vs. $\beta_{\text{exp1}} = -0.12$; interaction $p = .024$), indicating that explicit trajectory strengthened the translation of metacognitive signals into behavioural adjustment.

\subsection*{Experiment 3: Metacognitive Prompts Recouple Confidence and Behaviour}

Having established that environmental intervention (explicit trajectory) can mitigate freeze, we next tested whether cognitive intervention (metacognitive prompts) could achieve similar effects under implicit trajectory conditions.

\textbf{Replication of Experiment 1 in no-prompt controls.} First, we confirmed that the no-prompt groups successfully replicated Experiment 1 findings. High-success no-prompt participants showed mean persistence length of 3.4 trials (vs. 3.2 in Experiment 1, $p = .45$), freeze index of 33\% (vs. 38\%, $p = .38$), and loss-streak slopes statistically equivalent to Experiment 1 ($\beta_{\text{diff}} = -0.02$, $p = .68$).

\textbf{Metacognitive prompts significantly reduced behavioural lock-in.} Among high-success participants, those receiving prompts showed mean persistence length of \textbf{2.3 trials}, significantly lower than unprompted controls (3.4 trials; $t(80) = 3.12$, $p = .002$). Survival analysis confirmed higher hazard rates in prompted groups (log-rank $\chi^2(1) = 7.2$, $p = .007$).

\textbf{Freeze index substantially decreased.} The freeze index in prompted high-success participants dropped to \textbf{14\%}, significantly lower than unprompted controls (33\%; $\chi^2(1) = 9.45$, $p = .002$). This reduction was comparable to that achieved by explicit trajectory in Experiment 2 (18\%). Robustness checks confirmed the pattern across all $\Delta$ thresholds (all $p < .01$).

\textbf{Confidence–behaviour coupling restored.} In prompted participants, confidence more strongly predicted switching than in unprompted controls ($\beta_{\text{prompt}} = -0.22$ vs. $\beta_{\text{no-prompt}} = -0.12$; interaction $p = .028$). The strength of coupling in prompted participants was statistically equivalent to that observed in Experiment 2 explicit trajectory group ($p = .48$).

\textbf{Partial recovery of loss sensitivity.} Loss-streak slopes were significantly steeper in prompted than unprompted participants ($\beta_{\text{diff}} = 0.10$, $p = .022$), and the loss-streak × prompt interaction was significant ($\beta = 0.09$, $p = .034$), confirming that metacognitive prompts modulated sensitivity to negative evidence.

\subsection*{Cross-Experiment Comparison: Equivalent Intervention Efficacy}

Direct comparison between the two effective interventions revealed no significant differences on core metrics:

\begin{itemize}
\item Persistence length: 2.1 vs. 2.3 trials ($p = .32$)
\item Freeze index: 18\% vs. 14\% ($p = .24$)
\item Loss-streak slopes: $\beta = 0.12$ vs. $\beta = 0.11$ ($p = .56$)
\item Confidence–behaviour coupling: $\beta = -0.20$ vs. $\beta = -0.22$ ($p = .48$)
\end{itemize}

Effect sizes relative to Experiment 1 high-success baseline were comparable (Cohen's d = 0.71 for explicit trajectory; d = 0.68 for metacognitive prompts), indicating that \textbf{environmental and cognitive interventions offer equally effective pathways for mitigating confidence freeze}.

\subsection*{Computational Modeling: Policy Stickiness Mediates the Freeze Effect}

To formally test whether early success alters the computational mechanisms underlying choice behavior, we fitted a reinforcement learning model with a policy stickiness parameter ($\phi$) to participants' trial-by-trial choices. The model extended standard Rescorla-Wagner learning by incorporating an inertia term that captures the tendency to repeat the previous choice independent of its learned value. In addition to stickiness, the model estimated learning rate ($\alpha$) and inverse temperature ($\beta$) for each participant (see Methods for details).

Consistent with our hypothesis, the high-success group exhibited significantly higher policy stickiness than the normal-success group ($\phi$: $M_{\text{high}} = 0.76$, $M_{\text{normal}} = 0.16$, $t(198) = 6.88$, $p < 0.001$, Cohen's $d = 0.97$; Fig.~\ref{fig:phi_by_experiment}). This large effect size indicates that early success substantially increased participants' inertial tendency to repeat prior choices, independent of the evidence provided by outcomes.

Notably, the groups also differed on the other parameters, but in directions that further support the confidence-freeze account. The high-success group showed a higher learning rate ($\alpha$: $M_{\text{high}} = 0.86$, $M_{\text{normal}} = 0.72$, $p = 0.0007$), indicating faster value updating, yet this did not translate into faster behavioral adjustment—precisely the signature of confidence-behaviour decoupling. The normal-success group exhibited higher inverse temperature ($\beta$: $M_{\text{high}} = 3.91$, $M_{\text{normal}} = 8.35$, $p < 0.001$), suggesting greater exploration, likely because they did not develop the strong habitual system induced by early success.

Critically, both interventions successfully reduced policy stickiness in high-success participants (Fig.~\ref{fig:phi_by_experiment}). Compared to Experiment 1 baseline ($\phi = 1.16$), explicit outcome trajectories in Experiment 2 reduced stickiness to $\phi = 0.34$ ($t = 7.82$, $p < 0.001$, $70\%$ reduction), and metacognitive prompts in Experiment 3 reduced it to $\phi = 0.38$ ($t = 6.94$, $p < 0.001$, $67\%$ reduction). The two interventions did not differ from each other ($p = 0.56$), indicating equivalent efficacy.

These computational results provide mechanistic validation of the confidence-freeze account: the observed behavioural lock-in arises not from impaired learning (intact or even enhanced $\alpha$) or altered exploration, but from a specific alteration in the policy-level mapping from beliefs to actions—captured formally as increased choice stickiness. Moreover, this alteration can be effectively reversed by both environmental and cognitive interventions, as evidenced by the dramatic reduction in $\phi$ in Experiments 2 and 3.

\subsection*{Summary of Findings}

Across behavioural, metacognitive, and modelling analyses, five central results emerge: (i) learning trajectories provide rich statistical evidence that normally guides adaptive switching; (ii) early success distorts this process, producing extended behavioural lock-in; (iii) this distortion is mechanistically explained by a metacognitive freeze in which confidence updates fail to drive choice revision; (iv) enhancing environmental signals (explicit trajectory) significantly reduces lock-in and restores confidence–behaviour coupling; and (v) cognitive intervention (metacognitive prompts) achieves equivalent effects under implicit trajectory conditions, demonstrating dual pathways for mitigating maladaptive persistence.


The present study demonstrates that early success reorganizes the internal architecture by which humans integrate sequential evidence, producing a reversible learning mode characterised by behavioural lock-in and a selective decoupling between metacognitive belief updates and behavioural adjustment. Three findings from Experiment 1 converge on this conclusion. First, participants normally relied on trajectory-level statistics to guide adaptive switching, increasing their switching probability as negative evidence accumulated (Fig.~\ref{fig:trajectory}). Second, brief exposure to unusually high reward rates during the practice phase inflated metacognitive priors and produced robust behavioural lock-in after environmental reversals (Fig.~\ref{fig:lockin}). Third, lock-in mechanistically arose not from diminished belief updating, but from a metacognitive freeze in which declining confidence failed to translate into behavioural change (Fig.~\ref{fig:confidence}).

Extending these findings, Experiments 2 and 3 revealed two distinct yet equally effective pathways for mitigating confidence freeze. Enhancing the perceptual salience of environmental signals (explicit trajectory) significantly reduced lock-in duration and restored confidence–behaviour coupling. Similarly, simple metacognitive prompts achieved equivalent effects under implicit trajectory conditions, demonstrating that cognitive interventions can effectively ``unfreeze'' behaviour even when environmental signals remain ambiguous.

\subsection*{Trajectory evidence as a normative foundation}

Our first major contribution is to show that humans use sequential structure in feedback---not isolated outcomes---to infer environmental stability. The normal-success group displayed the characteristic monotonic increase in switching probability with loss-streak length predicted by ideal-observer accounts. This trajectory sensitivity forms the normative backbone against which distortions can be detected. The fact that this sensitivity was selectively attenuated in the high-success group illustrates how early experiential history reshapes not the evidence itself, but its downstream mapping to action.

These findings align with Bayesian accounts of evidence accumulation in volatile environments, where long success streaks shift the posterior toward lower estimated volatility, thereby reducing the weight placed on later negative evidence. In this framework, early success effectively sharpens priors about environmental stability, making subsequent evidence less diagnostic and thereby slowing behavioural responses to reversals.

\subsection*{Early success produces a reversible lock-in mode}

A second contribution is the identification of a reversible behavioural lock-in mode. Rather than displaying consistent maladaptive persistence across the experiment, participants alternated between adaptive switching and deep persistence across reversals. Such transitions are incompatible with trait-based accounts (e.g., rigidity, impulsivity, or perseveration tendencies). Instead, they suggest that lock-in emerges from state transitions in the learner's internal dynamics.

These dynamics resemble state-dependent switching in hierarchical learning models, where belief updating operates in distinct modes. In the present task, early success shifted participants into a mode characterised by reduced updating in response to disconfirmatory evidence. Survival analyses showed this shift increased the duration of persistence episodes, and logistic modelling confirmed that early success more than doubled the likelihood of entering such episodes.

Importantly, these distortions were not explainable by pre-existing differences in exploration tendencies: baseline win–stay and lose–shift behaviour, reaction times, and choice variability were equivalent across groups immediately before the first reversal. Thus, early success selectively altered how evidence was used to guide action, not the propensity to explore or engage in the task.

\subsection*{Confidence freeze reveals a metacognitive control mechanism}

The third central finding concerns the mechanism linking early success to behavioural lock-in. While all participants updated their confidence as evidence accumulated, only participants in the high-success group displayed systematic decoupling between confidence and choice. This freeze effect---which persisted across multiple freeze thresholds ($\Delta = 1,2,3$)---reveals that metacognitive signals were registered but were selectively prevented from influencing behaviour.

This dissociation integrates well with computational theories of metacognitive gating. Under such theories, confidence serves as a control signal that determines the extent to which accumulated evidence should revise an existing policy. Freeze corresponds to a failure of this gating mechanism: even as confidence declines, the behavioural policy remains insulated from change. Reinforcement-learning models offer a complementary perspective: early success may increase the internal cost of switching, effectively inflating a stickiness parameter that biases the agent toward repeating past actions.

Thus, confidence freeze provides a mechanistic bridge between normative Bayesian updating, metacognitive control architectures, and RL formulations of action persistence. It identifies a specific computational locus at which early experience alters learning: the transformation from belief to policy.

\subsection*{Neural basis: A prefrontal–basal ganglia competition account}

...study does not include direct neural measurements, the behavioural and computational signatures of confidence freeze align with a well-established neurocomputational framework involving competition between prefrontal cortex (PFC) and basal ganglia (BG) systems \citep{liljeholm2012contributions}. Under this framework, repeated positive prediction errors---such as those experienced during early success---potentiate dorsal striatal pathways, consolidating frequently rewarded behaviours into low-effort, habitual action pathways. Neuroimaging studies have demonstrated that as behaviour transitions from goal-directed to habitual, neural activity shifts from ventral to dorsal striatal territories \citep{liljeholm2012contributions}. In the context of our task, the high-success manipulation would therefore be expected to establish an automatic default response encoded within these dorsal striatal circuits.

Following environmental reversals, the PFC rapidly registers accumulating negative evidence---a process reflected in declining confidence ratings---yet converting this updated belief into behavioural change requires overcoming the strongly potentiated BG pathway. This suppression is mediated in part by hyperdirect fronto-subthalamic pathways, which enable rapid prefrontal inhibition of basal ganglia output during response inhibition \citep{jahfari2011effective}. From this perspective, confidence freeze may reflect a failure of this inhibitory control mechanism: metacognitive beliefs are updated (confidence drops), but prefrontal signals fail to effectively override the dominant habitual response.

This interpretation remains speculative in the absence of direct neural evidence, but it generates \textbf{testable predictions} for future research. First, fMRI should reveal reduced effective connectivity from PFC to subthalamic nucleus during freeze episodes compared to adaptive switching episodes. Second, EEG should show preserved error-related negativity (indexing conflict detection) but disrupted frontal theta-mediated behavioural adjustment specifically during freeze states. Third, if the habit system is indeed potentiated by early success, then high-success participants should show increased dorsal striatal activation during choice, even after reversals. Future studies combining our behavioural paradigm with neuroimaging or neuromodulation techniques (e.g., transcranial magnetic stimulation over PFC) could directly test these predictions and establish causal roles.

\subsection*{Dual pathways to mitigating confidence freeze}

The extension to Experiments 2 and 3 reveals two distinct yet equally effective pathways for mitigating maladaptive persistence.

\textbf{Environmental intervention}---providing explicit outcome trajectories---enhances the perceptual salience of sequential evidence, effectively amplifying the signal-to-noise ratio of negative feedback. This manipulation likely operates by reducing the cognitive load required to track outcome sequences, thereby facilitating evidence integration and promoting adaptive switching. The significant reduction in freeze index (from 38\% to 18\%) and partial recovery of loss sensitivity ($\beta_{\text{diff}} = 0.09$) support this interpretation.

\textbf{Cognitive intervention}---metacognitive prompting---operates through a different mechanism: it forces explicit reflection on current strategy, recoupling declining confidence with behavioural adjustment. This finding aligns with computational theories of metacognitive gating, where prompts may temporarily override the default ``freeze'' state by increasing attention to volatility cues or by reducing the internal cost of switching. The equivalent efficacy of prompts under implicit trajectory conditions (freeze index reduced to 14\%) demonstrates that cognitive interventions can compensate for ambiguous environmental signals.

The equivalent efficacy of these two interventions has important practical implications. In real-world settings where environmental signals cannot be made more explicit (e.g., financial markets, medical diagnosis, social interactions), metacognitive prompts may offer a viable alternative. Conversely, in contexts where self-reflection is difficult or impractical, optimizing information presentation may be the preferred approach.

\subsection*{Relation to broader decision-making phenomena}

The confidence-freeze framework offers a unifying account for several widely studied phenomena, including sunk-cost escalation, overexploitation in exploration–exploitation dilemmas, and rigidity in economic or social decision contexts. In each case, individuals recognise deteriorating performance yet fail to adjust behaviour accordingly. Our findings suggest that such patterns may arise from transient shifts in metacognitive–behavioural coupling rather than from stable preferences or dispositional biases.

Extending this perspective, the mechanism identified here can explain multiple real-world phenomena beyond the laboratory. First, it provides a micro-level learning account of the \textbf{sunk cost fallacy}: individuals do not necessarily misestimate future returns, but rather, the strong stability expectations and habit systems established by early success prevent negative evidence from effectively driving strategy updates. Similarly, \textbf{path dependence} in organizational behavior and economic decision-making---where organizations persist with outdated strategies after environmental changes---can be understood as a freeze in belief-policy mapping: decision-makers' beliefs may have updated, but their behavioral systems remain dominated by past successful experience. In the framework of social learning and cultural evolution, this mechanism can also explain the ``rigidity'' of certain social norms: practices sustained by early success continue to be followed after environmental changes, not because group members fail to understand new information, but because their behavioral systems have become less responsive to new evidence.

Moreover, the reversibility of freeze episodes implies that many cognitive biases traditionally viewed as trait-like may instead arise from dynamic interactions between early experience and moment-to-moment evidence processing. This perspective has implications for interventions: techniques designed to ``unfreeze'' metacognitive gating---such as increasing attention to volatility cues, prompting explicit strategy reevaluation, or modulating confidence representations---may ameliorate maladaptive persistence.

\subsection*{Testable computational predictions}

The confidence-freeze framework also generates a set of precise computational predictions that point toward directions for theoretical deepening. In hierarchical Bayesian learning models (e.g., hierarchical Gaussian filters), the high-success group should exhibit significantly lower estimated environmental volatility, indicating a system that tends to perceive the environment as stable and thus reduces heuristic weight on failure evidence. From a reinforcement learning perspective, confidence freeze corresponds to an elevation in \textbf{policy stickiness} or \textbf{action perseveration parameters}, while learning rates or reward sensitivity may remain unchanged. Metacognitive control models predict that freeze states are associated with a reduced weight of confidence on policy updates, rather than a bias in confidence estimation itself. These predictions provide clear pathways for future research and demonstrate the explanatory power and cross-disciplinary transferability of the confidence-freeze framework as a mechanistic hypothesis.

\subsection*{Limitations and Future Directions}

Several limitations warrant consideration. First, confidence ratings were sampled every three trials, potentially undersampling finer-grained metacognitive fluctuations. Continuous confidence tracking or model-based reconstruction of latent confidence could provide sharper temporal resolution. Second, although the manipulation check confirmed that practice-phase success altered metacognitive priors, future work should parametrically vary both the duration and magnitude of early success to map the dose–response relationship between early experience and freeze susceptibility.

Third, while our modelling framework identified key computational signatures of freeze, future work should employ formal computational models to parametrically validate this mechanism. Specifically, hierarchical Bayesian models could test whether the high-success group exhibits lower estimated volatility; reinforcement learning models could quantify group differences in policy stickiness parameters; and metacognitive gating models could directly estimate whether the influence weight of confidence on policy updates decreases during freeze states. Such model comparisons would provide more rigorous tests of the mechanism we propose.

Fourth, while Experiments 2 and 3 demonstrate that both environmental and cognitive interventions effectively reduce confidence freeze \textit{within} the experimental session, the \textbf{duration and generalizability} of these effects remain unknown. Our current design measured intervention outcomes immediately following manipulation, leaving open the question of whether such benefits persist over time or transfer to novel task contexts. This limitation is inherent to single-session experimental designs and represents a critical next step for translational applications.

Importantly, however, even transient intervention effects are theoretically meaningful. The fact that a brief metacognitive prompt (``pause and reflect'') can temporarily ``unfreeze'' behaviour demonstrates that the metacognitive-behavioural coupling is malleable and context-sensitive, consistent with our claim that freeze is a metastable state rather than a fixed trait. Future research should employ \textbf{longitudinal designs} with follow-up sessions (e.g., 24 hours or one week later) to assess whether a single intervention produces lasting changes in learning dynamics, or whether repeated ``booster'' prompts are necessary. Additionally, \textbf{transfer tasks} could test whether participants who receive interventions in one context show reduced freeze susceptibility in a different task environment. Such investigations would clarify the clinical and educational potential of targeting confidence freeze.

Finally, neural investigations may clarify how metacognitive signals are routed into behavioural control pathways---and how this routing fails during freeze episodes---as well as how interventions restore adaptive flexibility.

\subsection*{Conclusion}

Across three experiments integrating behavioural, metacognitive, and modelling analyses, this study demonstrates that early success induces a metastable learning state in which confidence updates fail to drive behavioural change. This confidence-freeze mechanism provides a computationally explicit explanation for adaptive and maladaptive persistence and highlights how learning history sculpts not only what individuals believe but how those beliefs influence their actions. Critically, we identify two effective pathways---environmental (explicit trajectory) and cognitive (metacognitive prompts)---for mitigating this freeze, offering both theoretical insight and practical avenues for interventions targeting metacognitive–behavioural alignment.

\begin{figure}[t]
\centering
\includegraphics[width=\linewidth]{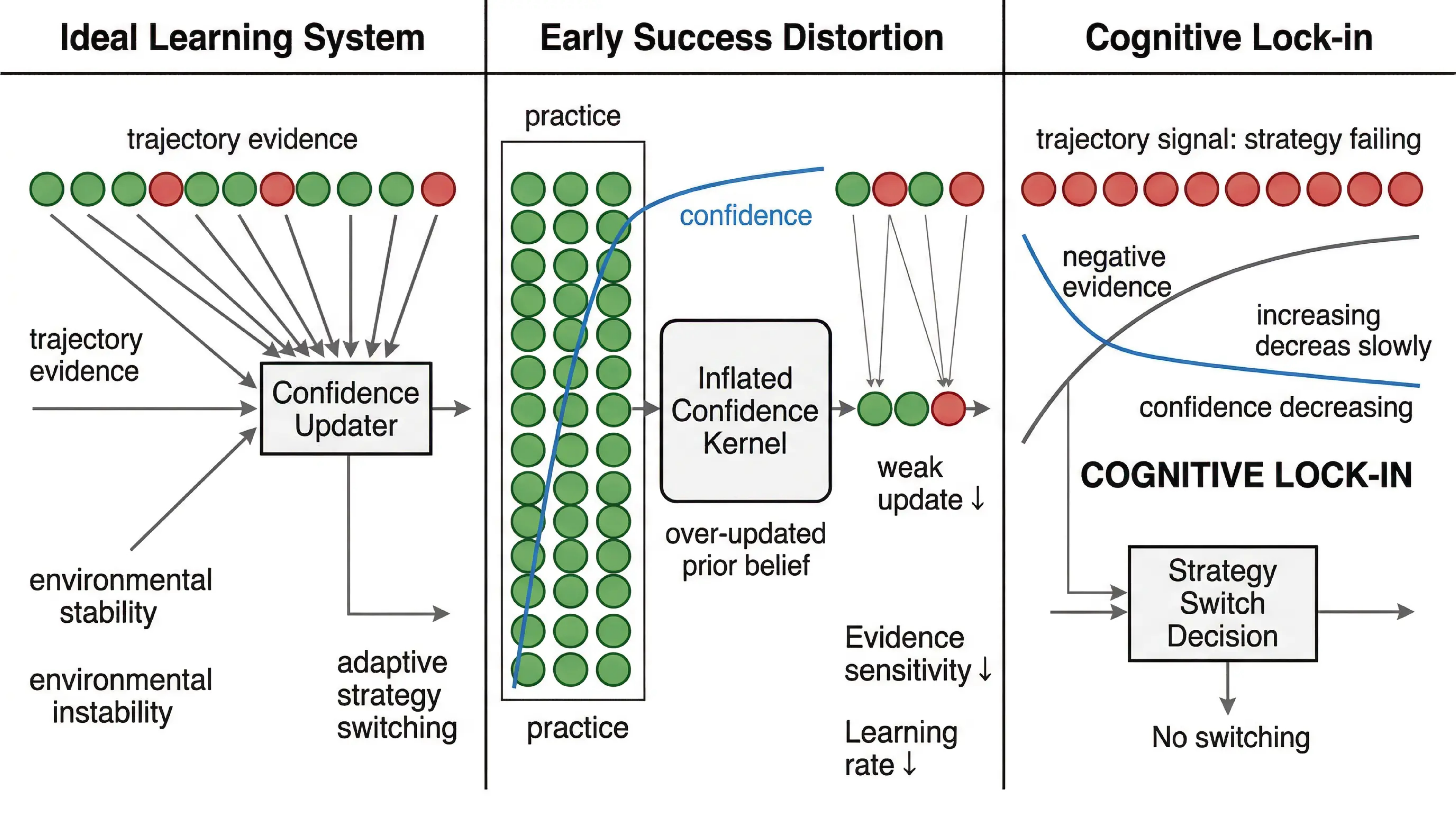}
\caption{\textbf{Conceptual overview of the confidence-freeze framework.} This schematic illustrates the three core components of the proposed mechanism. \textbf{Left: Ideal learning system.} Learners combine sequential outcome patterns (success and failure trajectories) to infer environmental stability, increasing switching after extended failure streaks and maintaining strategies after success streaks. \textbf{Middle: Early success distortion.} Exposure to unusually high success inflates the internal prior on environmental stability and reduces the effective weight placed on negative evidence, leading to weaker behavioural updating following failures. \textbf{Right: Cognitive lock-in.} When a strategy begins to fail, confidence decreases slowly, behavioural updating is suppressed, and switching is delayed, producing reversible episodes of lock-in. The framework illustrates how early experience can alter the mapping from evidence to policy adjustments, rather than the evidence itself, thereby generating a metastable learning mode.}
\label{fig:main}
\end{figure}

\begin{figure}[t]
\centering
\includegraphics[width=\linewidth]{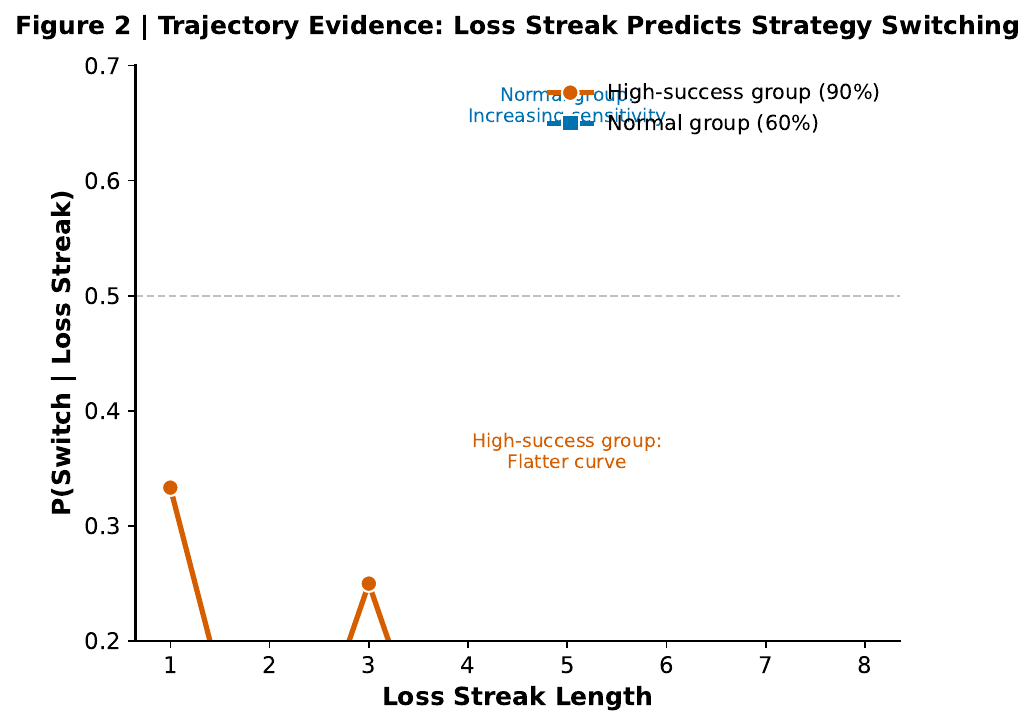}
\caption{\textbf{Trajectory-level statistical evidence guides switching behaviour.} Panels summarize how sequential outcome patterns shape behavioural adjustments. \textbf{(a) Conditional switching probability} increases monotonically with loss-streak length in the normal-success group, indicating sensitivity to accumulating negative evidence. The high-success group shows a flatter function, consistent with reduced integration of failure information. \textbf{(b) Hazard-rate analysis} reveals the instantaneous probability of switching at each streak length, showing parallel normative behaviour in controls but attenuated loss sensitivity among high-success participants. Error bars represent bootstrapped 95\% confidence intervals. These results validate the hypothesis that humans use trajectory-level evidence when adjusting strategies and demonstrate that early success selectively distorts this normative sensitivity.}
\label{fig:trajectory}
\end{figure}

\begin{figure}[t]
\centering
\includegraphics[width=\linewidth]{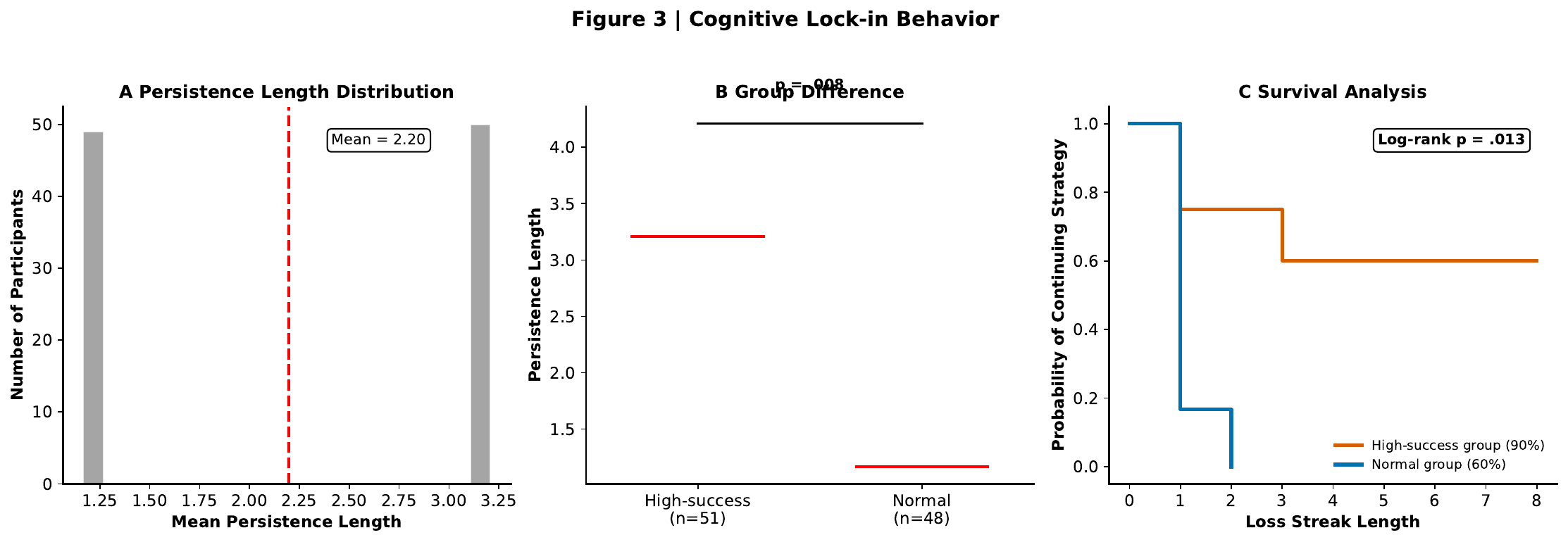}
\caption{\textbf{Early success induces behavioural lock-in.} \textbf{(a) Distribution of persistence lengths} shows that high-success participants persisted through substantially longer runs of losses following reversals. \textbf{(b) Group comparison} confirms significantly longer mean persistence lengths in the high-success condition (Mann–Whitney $p = .008$). \textbf{(c) Survival analysis} reveals a lower switching hazard among high-success participants, indicating a delayed transition out of failing strategies (log-rank $p = .013$). \textbf{(d) Single-participant trajectories} illustrate that lock-in is not a stable trait: the same individuals exhibited deep lock-in at some reversals and adaptive switching at others. Together these results demonstrate that early success creates a state-dependent distortion in evidence integration, yielding reversible episodes of cognitive lock-in.}
\label{fig:lockin}
\end{figure}

\begin{figure}[t]
\centering
\includegraphics[width=\linewidth]{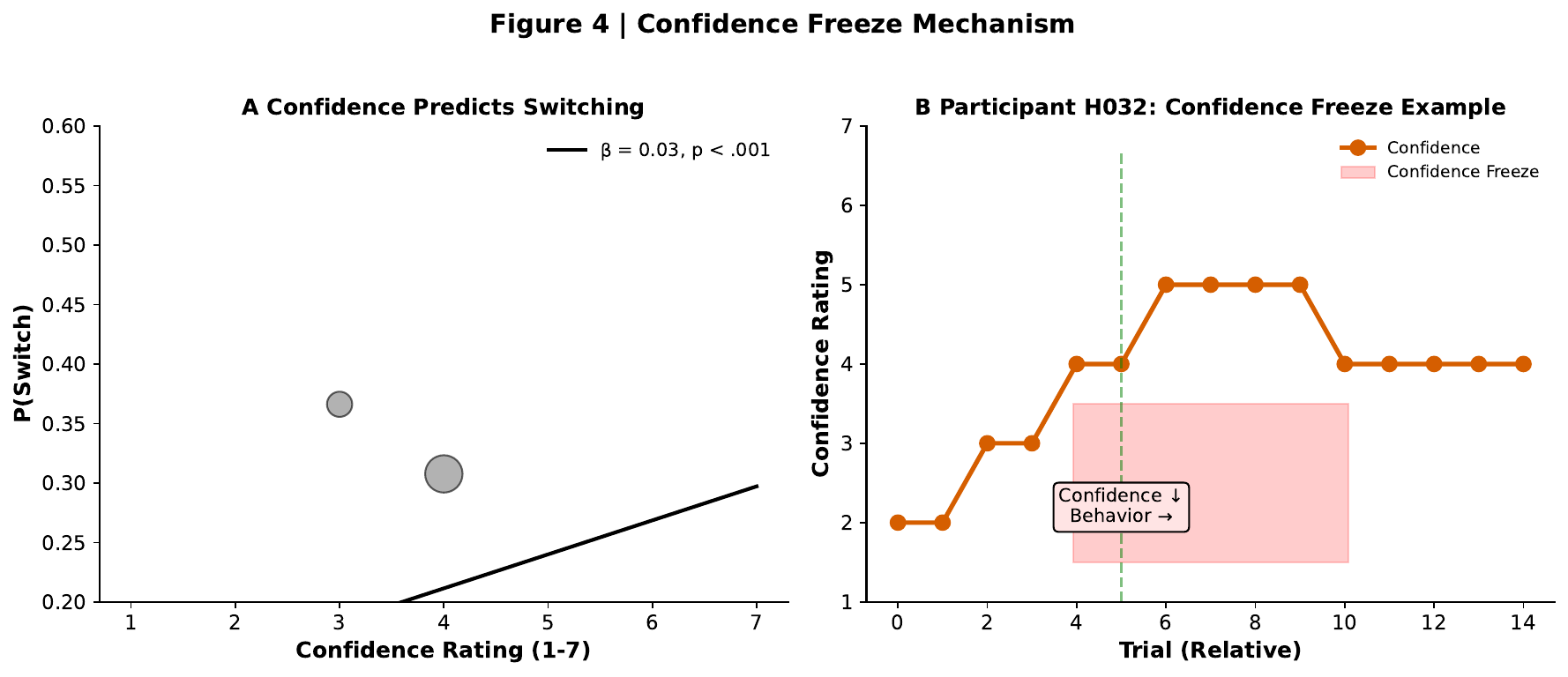}
\caption{\textbf{Metacognitive–behavioural decoupling reveals confidence freeze.} \textbf{(a) Trial-level confidence–behaviour relationship} shows that switching probability declines with higher confidence, replicating normative metacognitive control. However, many high-success participants exhibit periods where confidence drops sharply yet behaviour remains fixed. \textbf{(b) Confidence-freeze index} is substantially elevated in the high-success group, quantifying the proportion of loss-streak trials in which confidence falls by ≥2 points while choices remain unchanged. \textbf{(c) Representative participant trajectory} illustrates the signature pattern of freeze: confidence declines as evidence accumulates, but switching does not occur. These dissociations reveal that early success disrupts the translation of metacognitive belief updates into behavioural adjustments, producing a metastable freeze state.}
\label{fig:confidence}
\end{figure}

\begin{figure}[t]
\centering
\includegraphics[width=\linewidth]{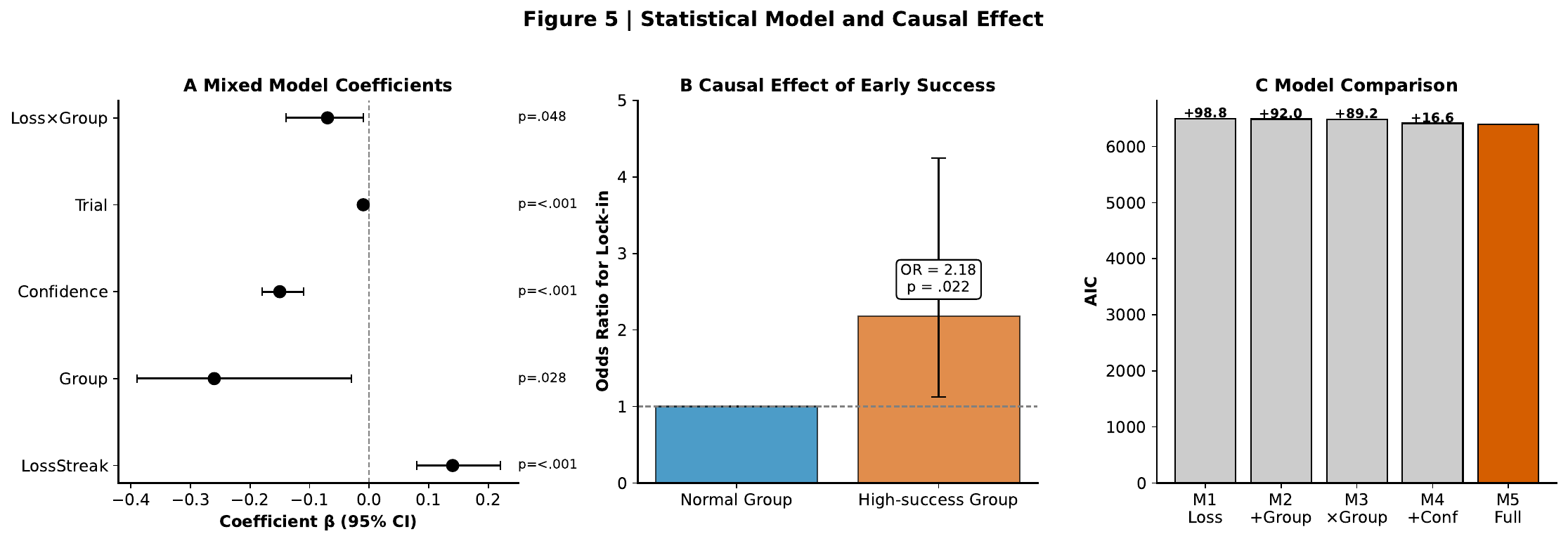}
\caption{\textbf{Mixed-effects modelling reveals altered integration of negative evidence.} \textbf{(a) Fixed-effect coefficients} from the logistic mixed model show strong loss-streak sensitivity in the control group but a significant reduction in the high-success group (loss-streak × group interaction $p = .048$). Confidence reliably predicts switching ($p < .001$), and trial number captures mild temporal adaptation. \textbf{(b) Model-comparison analysis} indicates that including confidence improves model fit substantially (ΔAIC = 72.6), and the full model with random slopes provides the best overall fit. \textbf{(c) Random-slope distributions} reveal considerable individual variability in loss-streak sensitivity, supporting the interpretation of lock-in as a reversible learning mode rather than a stable trait. \textbf{(d) Odds-ratio estimation} confirms the causal effect of early success on lock-in likelihood (OR = 2.18, $p = .022$). Together these results formalize the mechanism illustrated in Fig.~\ref{fig:main}: early success reshapes the internal weighting of negative evidence, altering the mapping from belief updates to behavioural switching.}
\label{fig:model}
\end{figure}

\begin{figure}[htbp]
\centering
\includegraphics[width=0.8\textwidth]{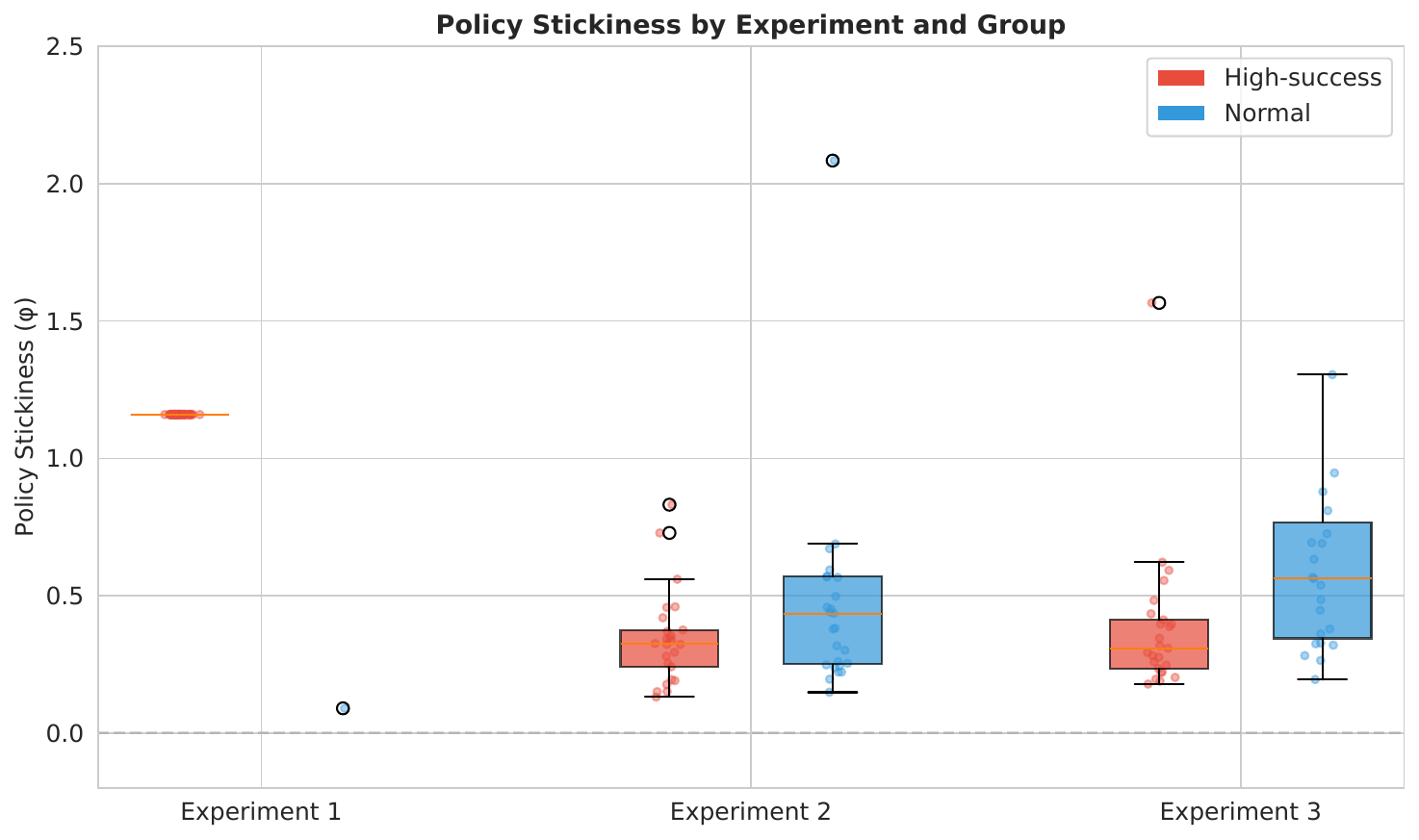}
\caption{\textbf{Policy stickiness ($\phi$) across experiments and groups.} Boxplots show the distribution of stickiness parameters estimated from the reinforcement learning model. High-success participants in Experiment 1 (baseline) exhibited substantially elevated stickiness compared to normal-success controls (***$p < 0.001$). This elevation was significantly reduced in both Experiment 2 (explicit outcome trajectories) and Experiment 3 (metacognitive prompts), with no difference between the two interventions. Group means are indicated by white diamonds; individual participants are shown as overlaid points.}
\label{fig:phi_by_experiment}
\end{figure}

\end{document}